\documentclass[runningheads]{llncs}
\usepackage{epsfig}
\usepackage{graphicx}
\usepackage{multirow}
\usepackage{amsmath,amssymb} 
\usepackage{color}

\begin{document}
\title{Dependency-aware Attention Control for Unconstrained Face Recognition with Image Sets} 

\titlerunning{Dependency-aware Attention Control with Image Sets}
%
\author{Xiaofeng Liu\inst{1}\orcidID{0000-0002-4514-2016} \and
B.V.K Vijaya Kumar \inst{1}\orcidID{0000-0001-7126-6381} \and
Chao Yang\inst{2}\orcidID{0000-0002-6553-7963}\and
Qingming Tang\inst{3}\orcidID{0000--0002-2670-4917}\and
Jane You\inst{4}\orcidID{0000-0002-8181-4836}}
%
\authorrunning{X. Liu, B.V.K Kumar, C. Yang, Q. Tang, J. You}
%

\institute{Carnegie Mellon University PA 15213, USA\\ \email{liuxiaofeng@cmu.edu} \and
University of Southern California CA 90089, USA
\\
\and
Toyota Technological Institute at Chicago IL 60637, USA\\\and
The Hong Kong Polytechnic University}
\maketitle              

\begin{abstract}
This paper targets the problem of image set-based face verification and identification. Unlike traditional single media (an image or video) setting, we encounter a set of heterogeneous contents containing orderless images and videos. The importance of each image is usually considered either equal or based on their independent quality assessment. How to model the relationship of orderless images within a set remains a challenge. We address this problem by formulating it as a Markov Decision Process (MDP) in the latent space. Specifically, we first present a dependency-aware attention control (DAC) network, which resorts to actor-critic reinforcement learning for sequential attention decision of each image embedding to fully exploit the rich correlation cues among the unordered images. Moreover, we introduce its sample-efficient variant with off-policy experience replay to speed up the learning process. The pose-guided representation scheme can further boost the performance at the extremes of the pose variation. 


\end{abstract}
\section{Introduction}

Recently, unconstrained face recognition (FR) has been rigorously researched in computer vision community \cite{chen2017unconstrained,learned2016labeled}. In its initial days, the single image setting is used for FR evaluations, $e.g.,$ Labeled Faces in the Wild (LFW) verification task \cite{huang2007labeled}. The trend of visual media explosion pushes the research into the next phase, where the video face verification attracts much attention, such as the YouTube Faces (YTF) dataset \cite{wolf2011face}. Since the LFW and YTF have a well-known frontal pose selection bias, the unconstrained FR is still considered an unsolved problem \cite{phillips2015human,crosswhite2017template}. In addition, the open-set face identification is actually more challenging compared to the verification popularized by the LFW and YTF datasets \cite{hayat2017joint,liu2017sphereface}.

The IARPA Janus Benchmark A (IJB-A) \cite{klare2015pushing} provides a more practical unconstrained face verification and identification benchmark. It takes a set (containing orderless images and/or videos with extreme head rotations, complex expressions and illuminations) as the smallest unit for representation. The set of a subject can be sampled from the mugshot history of a criminal, lifetime enrollment images for identity documents, different check points, and trajectory of a face in the video. This kind of setting is more similar to the real-world biometric scenarios \cite{grother2014face}. Capturing human faces from multiple views, background environments, camera parameters, does result in the large inner-set variations, but also incorporates more complementary information hopefully leading to higher accuracy in practical applications \cite{liu2017quality}.

\begin{figure}[t]
\centering
\includegraphics[height=3.2cm]{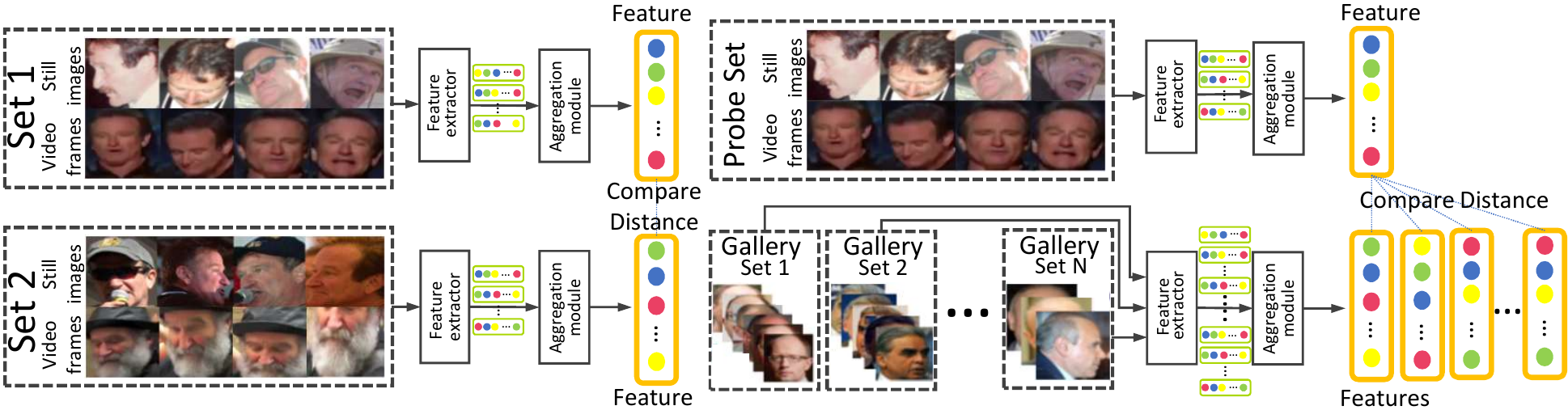}
\caption{ Illustration of the image set-based 1:1 face verification (left) and open-set 1:N identification (right) using the typical aggregation method.}
\label{fig:example}
\end{figure}

A commonly adopted strategy to aggregate identity information in each image is the average/max pooling \cite{li2014eigen,parkhi2015deep,chen2015end,chowdhury2016one}. Since the images vary in quality, a neural network-based assessment module has been introduced to independently assign the weight for each frame \cite{liu2017quality,yang2017neural}. By doing this, the frontal and clear faces are favored by their model. However, this may result in redundancy and sacrifice the diversity in a set. As shown in Fig. 2, these inferior frontal images are given relatively high weights in a set, sometimes as high as the weight given to the most discriminative one. There is little additional information that can be extracted from the blurry version of the same pose, while the valuable profile information $etc.,$ are almost ignored by the system. We argue that the desired weighting decision should depends on the other images within a set.
\begin{figure}[t!]
\centering
\begin{tabular}{cc}
\includegraphics[height=3.8cm]{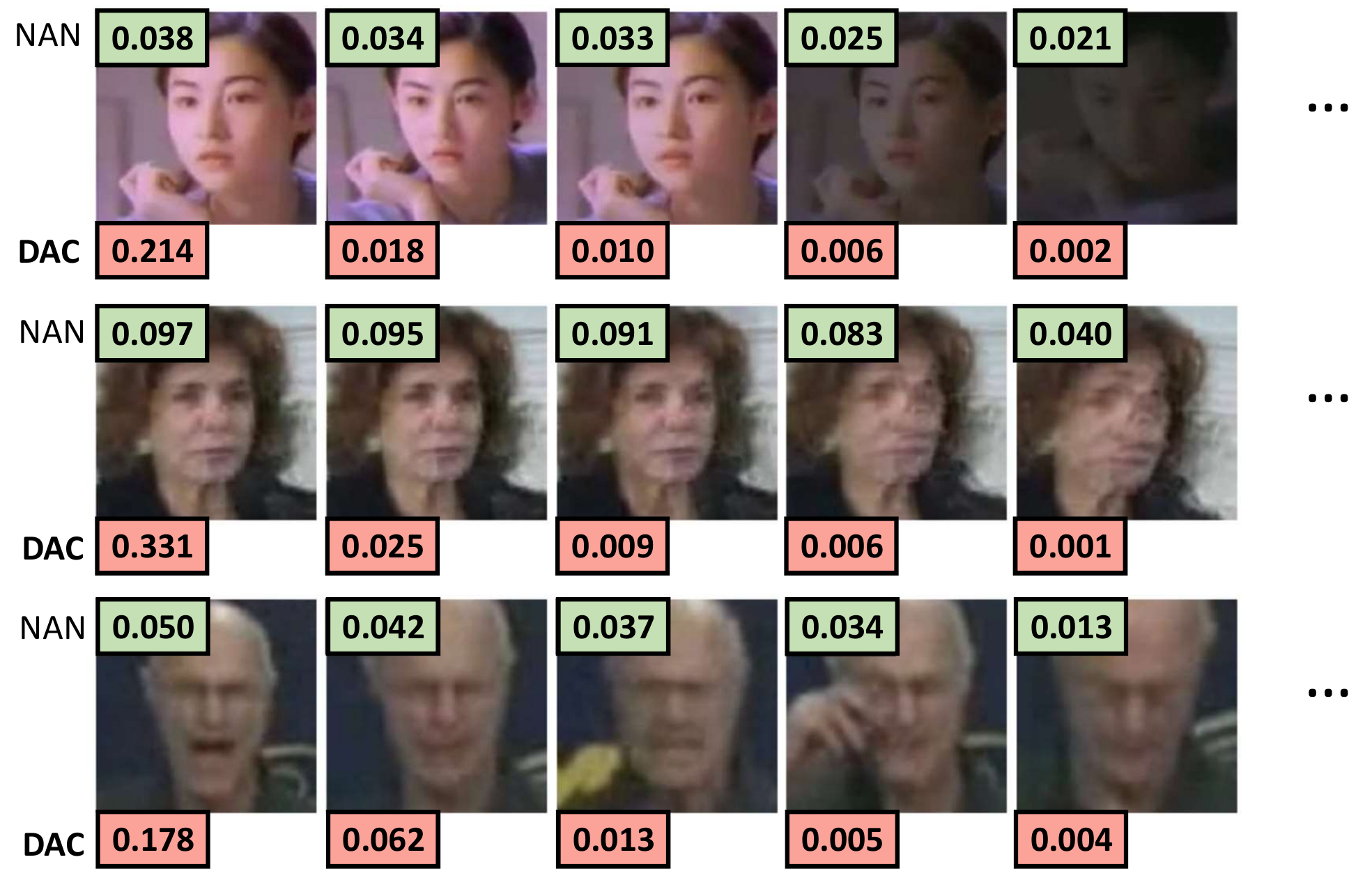}&\includegraphics[height=3.8cm]{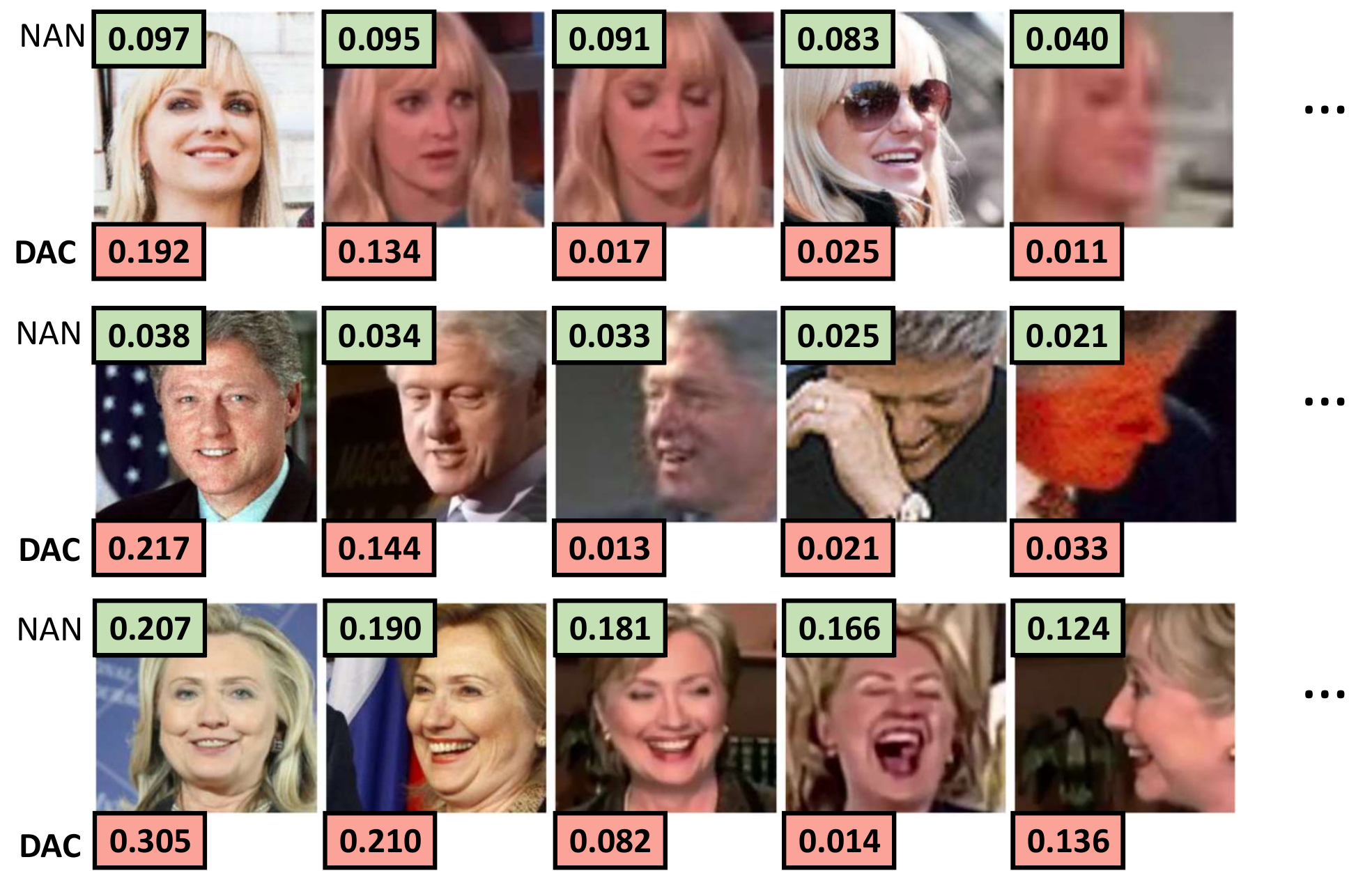}\\
(a)&~~~~~~(b)
\end{tabular}
\caption{Typical examples on the test set of (a) YTF and (b) IJB-A dataset showing the weights of images calculated by the previous method NAN \cite{yang2017neural}, and proposed DAC.}
\label{fig:example}
\end{figure}

Instead, we propose to formulate the attention scheme as a Markov Decision Process and resort to the actor-critic reinforcement learning (RL) to harness model learning. The dependency-aware attention control (DAC) module learns a policy to decide the importance of each image step-by-step with the observation of the other images in a set. In this way, we adaptively aggregate the feature vectors into a highly-compact representation inside the convex hull spanned by them. It not only explicitly learns to advocate high-quality images while repelling low-quality one, but also considers the inner-set dependency to reduce the redundancy and maintains the benefit of diversity information.

Moreover, extracting a set-level invariable feature can be always challenging to incorporate all of the potential information in varying poses, illumination conditions, resolutions $etc.$ Some approaches aggregate the image-level pair-wise similarity scores of two compared sets to fully use all images \cite{taigman2014deepface,schroff2015facenet,sun2015deeply,wen2016discriminative}. Given $n$ as the average number of images in a set, then this corresponds to the $\mathcal{O}(n^2)$ computational complexity per match operation and $\mathcal{O}(n)$ space complexity per set are not desirable. More recently, \cite{zhang2017multi,rao2017attention,janisch2017classification} are proposed to trade-off between speed and accuracy of processing paired video inputs using value-based Q-learning methods. These configurations focus on the verification, and cannot be scaled well for large scale identification tasks \cite{yang2017neural}. Conventionally, the feature extraction of the probe and gallery samples are independent processes \cite{liu2017sphereface}.

We notice that pose is the primary challenge in the IJB-A dataset and real applications \cite{hayat2017joint,zhu2014multi,wright2009robust}, and there is a prior that the structures of frontal and profile face are significantly different. Therefore, we simply utilize a pose-guided representation (PGR) scheme with stochastic routing to model the inter-set dependency. It well balances the computation cost and information utilization. 

Considering the above factors, we propose to fully exploit both the inner- and inter-set relationship for the unified set-based face verification and identification. (1) To the best of our knowledge, this is the first effort to introduce deep actor-critic RL into visual recognition problem. (2) The DAC can potentially be a general solution to incorporate rich correlation cues among orderless images. Its coefficients can be trained in a normal recognition training task given only set-level identity annotation, without the need of extra supervision signals. (3) To further improve the sample-efficiency, the trust region-based experience replay is introduced to speedup the training and achieve a stronger convergence property. (4) The PGR scheme well balances the computation cost and information utilization at the extremes of pose variation with the prior domain knowledge of human face. (5) The module-based feature-level aggregation also inherits the advantage of conventional pooling strategies $e.g.,$ taking varied number of inputs as well as offering time and memory efficiency.

We show that our method leads to the state-of-the-art accuracy on IJB-A dataset and also generalizes well in several video-based face recognition tasks, \textit{e.g.}, YTF and Celebrity-1000.

\section{Related work}

\noindent{\bf{Image set/video-based face recognition}} has been actively studied in recent years \cite{learned2016labeled}. The multi-image setting in the template-based dataset is similar to the multiple frames in the video-base recognition task. However, the temporal structure within a set is usually disordered, and the inner/inter-set variations are more challenging \cite{klare2015pushing}. We will not cover the methods which exploit the temporal dynamics here. There are two kinds of conventional solutions, $i.e.,$ manifold-based and image-based methods. In the first category, each set/video is usually modeled as a manifold, and the similarity or distance is measured in the manifold-level. In previous works, the affine hull, SPD model, Grassmann manifolds, $n$-order statistics and hyperplane similarity have been proposed to describe the manifolds \cite{cevikalp2010face,huang2016building,huang2017riemannian,wang2012covariance,lu2013image}. In these methods, images are considered as equal importance. They usually cannot handle the large appearance variations in the unconstrained FR task. For the second category, the pairwise similarities between probe and gallery images are exploited for verification \cite{taigman2014deepface,schroff2015facenet,wen2016discriminative,sivic2009you,lu2015multi}. The quadratic number of comparisons make them not scale well for identification tasks. Yang $et~al.$ \cite{yang2017neural} propose an attention model to aggregate a set of features to a single representation with an independent quality assessment module for each feature. Reference \cite{rao2017learning} further up-sampled the aggregated features to an image, then fed it to an image-based FR network. However, weighting decision for an image does not take the other images into account as discussed in Sec. 1. Since the frequently used RNN in video task \cite{graves2014neural,vinyals2015order,rao2017attention} is not fit for the image set, in this work, we consider the dependency within a set of features in a different way, where we use deep reinforcement learning to suggest the attention of each feature.

\noindent{\bf{Reinforcement learning (RL)}} trains an agent to interact (by trail and error) with a dynamic environment with the objective to maximize its accumulated reward. Recently, deep RL with convolutional neural networks (CNN) achieved human-level performance in Atari Games \cite{mnih2015human}. The CNN is an ideal approximate function to address the infinite state space \cite{li2017deep}. There are two main streams to solve RL problems: methods based on value function and methods based on policy gradient. The first category, $e.g.,$ Q-learning, is the common solution for discrete action tasks \cite{mnih2015human}. The second category can be efficient for continuous action space \cite{silver2014deterministic,lillicrap2015continuous}. There is also a hybrid actor-critic approach in which the parameterized policy is called an actor, and the learned value-function is called a critic \cite{mnih2016asynchronous,babaeizadeh2017reinforcement}. As it is essentially a policy gradient method, it can also be used for continuous action space \cite{arulkumaran2017deep}.

Besides, policy-based and actor-critic methods have faster convergence characteristics than value-based methods \cite{sutton2000policy}, but they usually suffer from low sample-efficiency, high variance and often converge to local optima, since they typically learn via on-policy algorithms \cite{williams1992simple,schulman2016high}. Even the Asynchronous Advantage Actor-Critic \cite{mnih2016asynchronous,babaeizadeh2017reinforcement} also requires new samples to be collected for each gradient step on the policy. This quickly becomes extravagantly expensive, as the number of gradient steps to learn an effective policy increases with task complexity. Off-policy learning instead aims to reuse past experiences. This is not directly feasible with conventional policy gradient formulations, despite it relatively straightforward for value-based methods \cite{li2017deep}. Hence in this paper, we focus on combining the stability of actor-critic methods with the efficiency of off-policy RL, which capitalizes in recent advances on deep RL \cite{mnih2016asynchronous}, especially off-policy algorithms \cite{schulman2015trust,wang2017sample}. 

In addition to its traditional applications in robotics and control, recently RL has been successfully applied to a few visual recognition tasks. Mnih \textit{et al.} \cite{mnih2014recurrent} introduce the recurrent attention model to focus on selected regions or locations from an image for digits detection and classification. This idea is extended to identity alignment by iteratively removing irrelevant pixels in each image \cite{lan2017identity}. The value-based Q-learning methods are used for object tracking \cite{huang2017learning} and the video verification in a computationally efficient view by dropping inefficient probe-gallery pairs \cite{rao2017attention} or stopping the comparison after receiving sufficient pairs \cite{zhang2017multi,janisch2017classification} . However, this will inevitably result in information loss of the unused pairs and only applicable for verification. There has been little progress made in policy gradient/actor-critic RL for visual recognition.

\section{Proposed methods}
The flow chart of our framework is illustrated in Fig. 3. It takes a set of face images as input and processes them with two major modules to output a single(w/o PGR)/three(with PGR) feature vectors as its representation for recognition. We adopt a modern CNN module to embed an image into a latent space, which can largely reduce the computation costs and offer a practicable state space for RL. Then, we cascade the DAC, which works as an attention model reads all feature vectors and linearly combines them with adaptive weighting at the feature-level. Following the memory attention mechanism described in \cite{graves2014neural,vinyals2015order,yang2017neural}, the features are treated as the memory and the feature weighting is cast as a memory addressing procedure.These two modules can be trained in a one-by-one or end-to-end manner. We choose the first option, which makes our system benefit from the sufficient training data of the image-based FR datasets. The PGR scheme can further utilize the prior knowledge of human face to address a set with large pose variants.

\begin{figure}[t!]
\centering
\includegraphics[height=6.2cm]{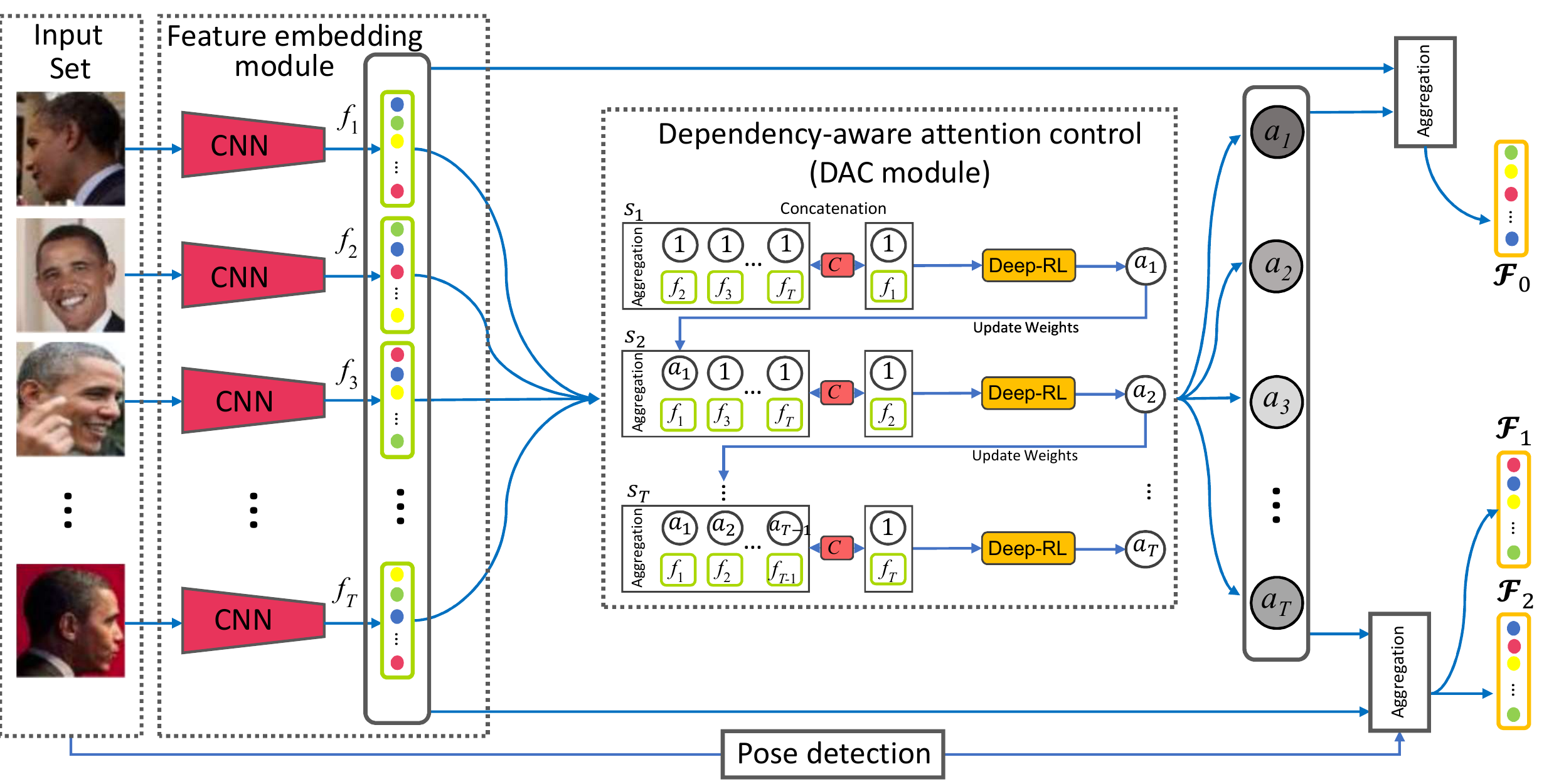}
\caption{ Our network architecture for image set-based face recognition.}
\label{fig:example}
\end{figure}

\subsection{Inner-set dependency control}

In the set-based recognition task, we are given $M$ sets/videos $\mathop{({\mathcal{X}}^m, y^m)}_{m=1}^{M}$, where ${\mathcal{X}}^m$ is a image set/video sequence with varying number of images $T^m$ (\textit{i.e.,} ${\mathcal{X}}^m=\left\{{\mathop{x}_{1}^{m}},{\mathop{x}_{2}^{m}},\cdots,{\mathop{x}_{T^m}^{m}}\right\}$, ${\mathop{x}_{t}^{m}}$ is the $t$-th image in a set) and the $y^m$ is the corresponding set-level identity label. We feed each image ${\mathop{x}_{t}^{m}}$ to our model, and its corresponding feature representation ${\mathop{f}_{t}^{m}}$ are extracted using our neural embedding network. Here, we adopt the GoogLeNet \cite{szegedy2016rethinking} with Batch Normalization \cite{ioffe2015batch} to produce a 128-dimensional feature as our encoding of each image. With a relatively simple architecture, GoogLeNet has shown superior performance on several FR benchmarks. It can be easily replaced by other advanced CNNs for better performance. In the rest of the paper, we will simply refer to our neural embedding network as CNN, and omit the upper index (identity) where appropriate for better readability.

Since the features are deterministically computed from the images, they also inherit and display large variations. Simply discarding some of them using the hard attention scheme may result in loss of too much information in a set \cite{yang2017neural,rao2017attention}. Our attention control can be seen as the task of reinforcement learning to find the optimal weights of soft attention, which defines how much of them are focused by the memory attention mechanism. Moreover, the principle of taking different number of images without temporal information, and having trainable parameters through standard recognition training are fully considered.

Our solution of inner-set dependency modeling is to formulated as a MDP. At each time step $t$, the agent receives a state ${s_t}$ in a state space $\mathcal{S}$ and chooses an action ${a_t}$ from an action space $\mathcal{A}$, following a policy $\pi(a_t\mid s_t)$, which is the behavior of the agent. Then the action will determine the next state \textit{i.e.,} $s_{t+1}$ or termination, and receive a reward ${r_t(s_t,a_t)}\in \mathcal{R} \subseteq \mathbb{R}$ from the environment. The goal is to find an optimal policy $\pi^*$ that maximizes the discounted total return $R_t=\sum_{i\geq0}^{T}\gamma^{i}{r_{t+i}(s_t,a_t)}$ in expectation, where $\gamma\in [0,1)$ is the discount factor to trade-off the importance of immediate and future rewards \cite{li2017deep}.

In the context of image-set based face recognition, we define the actions, $i.e., \left\{a^{1},a^{2},\cdots,a^{T}\right\}$, as the weights of each feature representation ${\mathop{\left\{f\right\}}_{i=1}^{T}}$.  The weights of soft attention ${\mathop{\left\{a\right\}}_{i=1}^{T}}$ are initialized to be 1, and are updated step-by-step. The state $s_t$ is related to the $t-1$ weighted features and $T-(t-1)$ to-be weighted features. In contrast to image-level dependency modeling, the compact embeddings largely shrink the state space and make our RL training feasible. In our practical applications, $s_t$ is the concatenation of $f_t$ and the aggregation of the remaining features with their updated weights at time step $t$. The termination means all of the images in this set have been successfully traversed.\begin{align}
{s_t}= \left\{\frac{(\sum_{i=1}^T {a_i}{f_i})-{f_t}}{(\sum_{i=1}^T {a_i})-1} \right\} Concatenate \left\{{f_t}\right\}
\end{align}

We define the global reward for RL by the overall recognition performance of the aggregated embeddings, which drives the RL network optimization. In practice, we add on the top of the DAC few fully connected layers $h$ and followed by a softmax to calculate the cross-entropy loss $L_m = -\log\left(\left\{e^{o_{y^m}}\right\}/{ \sum_j^M e^{o_j} }\right)$ to calculate the reward at this time step. We use the notation $o_j$ to mean the $j$-th element of the vector of class scores $o$. $g(\cdot)$ is the weighted average aggregation function, $h$ maps the aggregated feature with the updated weights $g({\mathcal{X}}^m\mid s_{t})$ to the $o$. The reward are defined as follows:\begin{align}
g({\mathcal{X}^m}\mid s_{t},\text{CNN})=\sum_{i=1}^{T^m} \frac{{a_i}{f_i}^m}{\sum{a_i}}~~(\text{with~updated}~a_i~\text{at~step}~t)
\end{align}\begin{align}
{r_t}= \left\{L_m[h(g({\mathcal{X}}^m\mid s_{t}))]-L_m[h(g({\mathcal{X}}^m\mid s_{t+1}))]\right\}+\lambda\text{max}[0,({1-a_t})]
\end{align}where the hinge loss term serves as a regularization to encourage redundancy elimination and is balanced with the $\lambda$. It also contributes to stabilize training. The aggregation operation essentially selects a point inside of the convex hull spanned by all feature vectors \cite{cevikalp2010face}.

Considering that the action space here is a continuous space $\mathcal{A}\in\mathbb{R}^+$, the value-based RL ($e.g.,$ Q-Learning) cannot tackle this task. We adapt the actor-critic network to directly grade each feature dependent on the observation of the other features. In a policy-based method, the training objective is to find a parametrized policy ${\pi}_\theta(a_t\mid s_t)$ that maximizes the expected reward $J(\theta)$ over all possible aggregation trajectories given a starting state. Following the Policy Gradient Theorem \cite{sutton2000policy}, the gradient of the parameters given the objective function has the form:\begin{align}
 {\nabla}_\theta J(\theta)=\mathbb{E}[{\nabla}_\theta \text{log} {\pi}_\theta(a_t\mid s_t)({Q}(s_t,a_t)-b(s_t))] 
\end{align}

\noindent where ${Q}(s_t,a_t)=\mathbb{E}[R_t\mid s_t,a_t]$ is the state-action value function, in which the initial action $a_t$ is provided to calculate the expected return when starting in the state $s_t$. A baseline function $b(s_t)$ is typically subtracted to reduce the variance while not changing the estimated gradient \cite{williams1992simple,andrew1999reinforcement}. A natural candidate for this baseline is the state only value function ${V}(s_t)=\mathbb{E}[R_t\mid s_t]$, which is similar to $Q(s_t,a_t)$, except the $a_t$ is not given here. The advantage function is defined as ${A}(s_t,a_t)={{Q}(s_t,a_t)}-{{V}(s_t)}$ \cite{li2017deep}. Eq.(4) then becomes:\begin{align}
 {\nabla}_\theta J(\theta)=\mathbb{E}[{\nabla}_\theta \text{log} {\pi}_\theta(a_t\mid s_t){A}(s_t,a_t)] 
\end{align} 

This can be viewed as a special case of the actor-critic model, where ${\pi}_\theta(a_t\mid s_t)$ is the actor and the ${A}(s_t,a_t)$ is the critic. To reduce the number of required parameters, the parameterized temporal difference (TD) error ${\delta_\omega}={r_t}+{\gamma}V_{\omega}(S_{s+1})-V_{\omega}(S_{s})$ can be used to approximate the advantage function \cite{schulman2016high}. We use two different symbols $\theta$ and $\omega$ to denote the actor and critic function, but most of these parameters are shared in a main stream neural network, then separated to two branches for policy and value predictions, respectively.

\subsection{Off-policy actor-critic with experience replay}

On-policy RL methods update the model with the samples collected via the current policy. The experience replay (ER) can be used to improve the sample-efficiency\cite{lin1992self}, where the experiences are randomly sampled from a replay pool $\mathcal{P} $. This ensure the training stability by reducing the data correlation. Since these past experiences were collected from different policies, the use of ER leads to off-policy updates. 

When training models with RL, $\varepsilon$-greedy action selection is often used to trade-off between exploitation and exploration, whereby a random action is chosen with a probability otherwise the top-ranking action is selected. A policy used to generate a training weight is referred to as a behavior policy $\mu$, in contrast to the policy to-be optimized which is called the target policy $\pi$.

The basic advantage actor-critic (A2C) training algorithm described in Sec. 3.1 is on-policy, as it assume the actions are drawn from the same policy as the target to-be optimized (i.e., $\mu=\pi$). However, the current policy $\pi$ is updated with the samples generated from old behavior policies $\mu$ in off-policy learning. Therefore, an importance sampling (IS) ratio is used to rescale each sampled reward to correct the sampling bias at time-step $t$: $\rho_t=\pi(a_t\mid s_t)/\mu(a_t\mid s_t)$ \cite{meuleau2000off}. For A2C, the off-policy gradient for the parametrized state only value function $V_\omega$ thus has the form:

\begin{align}
\Delta\omega^{\text{off}}=\sum_{t=1}^{T}(\bar{R}_t-\hat{V}_\omega(s_t))\nabla_\omega\hat{V}_\omega(s_t)\prod_{i=1}^t {\rho_i} 
\end{align}

\noindent where $\bar{R}_t$ is the off-policy Monte-Carlo return \cite{precup2001off}:

\begin{align}
\bar{R}_t=r_t+\gamma r_{t+1}\prod_{i=1}^1 {\rho_i}+\cdots+\gamma^{T-t} r_{T}\prod_{i=1}^{T-t} {\rho_{t+i}}
\end{align}

Likewise, the updated gradient for policy $\pi_\theta$ is:

\begin{align}
\Delta\theta^{\text{off}}=\sum_{t=1}^{T}\rho_{t} {\nabla}_\theta \text{log} {\pi}_\theta(a_t\mid s_t) \hat{\delta}_\omega
\end{align}

\noindent where $\hat{\delta}_\omega=r_t+\gamma \hat{V}_\omega(s_{t+1}-\hat{V}_\omega(s_{t})$ is the TD error using the estimated value of $\hat{V}_\omega$.

Here, we introduce a modified Trust Region Policy Optimization method \cite{schulman2015trust,wang2017sample}. In addition to maximizing the cumulative reward $J(\theta)$, the optimization is also subject to a Kullback-Leibler (KL) divergence limit between the updated policy $\theta$ and an average policy $\theta_a$ to ensure safety. This average policy represents a running average of past policies and constrains the updated policy from deviating too far from the average $\theta_a\leftarrow[\alpha \theta_a+(1-\alpha)\theta]$ with a weight $\alpha$. Thus, given the off-policy policy gradient $\Delta\theta^{\text{off}}$ in Eq.(8), the modified policy gradient with trust region $z$ is calculated as follows:

\begin{equation}
     \begin{aligned}
{\mathop{}_{z}^{minimize}}&~~\frac{1}{2}\|\Delta\theta^{\text{off}}-z\|_2^2,\\
\text{Subject to:}& \nabla_\theta D_{KL}[\pi_{\theta_a}(s_t)\|\pi_{\theta}(s_t)]^{\text{T}} ~z \leq\xi  
\end{aligned}
\end{equation}

\noindent where $\pi$ is the policy parametrized by $\theta$ or $\theta_a$,
and $ \xi$ controls the magnitude of the KL constraint. Since the constraint is linear, a closed form solution to this quadratic programming problem can be derived using the KKT conditions. Setting $k=\nabla_\theta D_{KL}[\pi_{\theta_a}(s_t)\|\pi_{\theta}(s_t)]$, we get:

\begin{align}
z_{tr}^*=\Delta\theta^{\text{off}}-max\left\{  \frac{k^{\text{T}}\Delta\theta^{\text{off}}-\xi}{\|k\|_2^2},0\right\}k
\end{align}

This direction is also shown to be closely related to the natural gradient \cite{amari1998natural,peters2006policy}. The above enhancements speed up and stabilize our A2C network training.

\subsection{Pose-guided inter-set dependency model}

To model the inter-set dependency without paired-input, we propose a pose-guided stochastic routing scheme. Such a divide-and-conquer idea originated in \cite{li2006bagging}, which constructs several face detectors to charge each view. Given a set of face image, we extract its general feature aggregation $\mathcal{F}_0$, as well as the aggregation of the frontal face features $\mathcal{F}_1$ and profile face feature $\mathcal{F}_2$. The $\mathcal{F}_1$ and $\mathcal{F}_2$ are the weighted average of the features from the near-frontal face images ($\leq 30^{\circ}$)  and profile face images ($>30^{\circ}$) respectively, in which the attention is assigned with the observation of full set. We use PIFA \cite{jourabloo2017pose} to estimate the yaw angle. The sum of weights of the frontal and profile features $p_1$ and $p_2$ are with respect to the quality of each pose group. Considering the mirror transforms in data augmentation and the symmetry property of human faces, we do not discriminate the right face and the left face. With PGR, the distance $d$ between two sets of samples is computed as:\begin{align}
d= \frac{1}{2}S(\mathcal{F}^1_0,\mathcal{F}^2_0)+\frac{1}{2}\sum_{i=1}^{2} {\sum_{j=1}^2 {S(\mathcal{F}^1_i,\mathcal{F}^2_j)p^1_ip^2_j}} 
\end{align}where $S$  is the L2 distance function to measure the distance between two feature vectors. We treat the generic features and pose-specific features equally, and fuse them for evaluations. The number of distance evaluations is decreased to $\mathcal{O}(5n)$.  This achieves promising verification performance requiring fewer comparisons than conventional image-level similarity measurements. It is also readily applied to the other variations.

\begin{figure}[t]
\centering
\includegraphics[height=3.7cm]{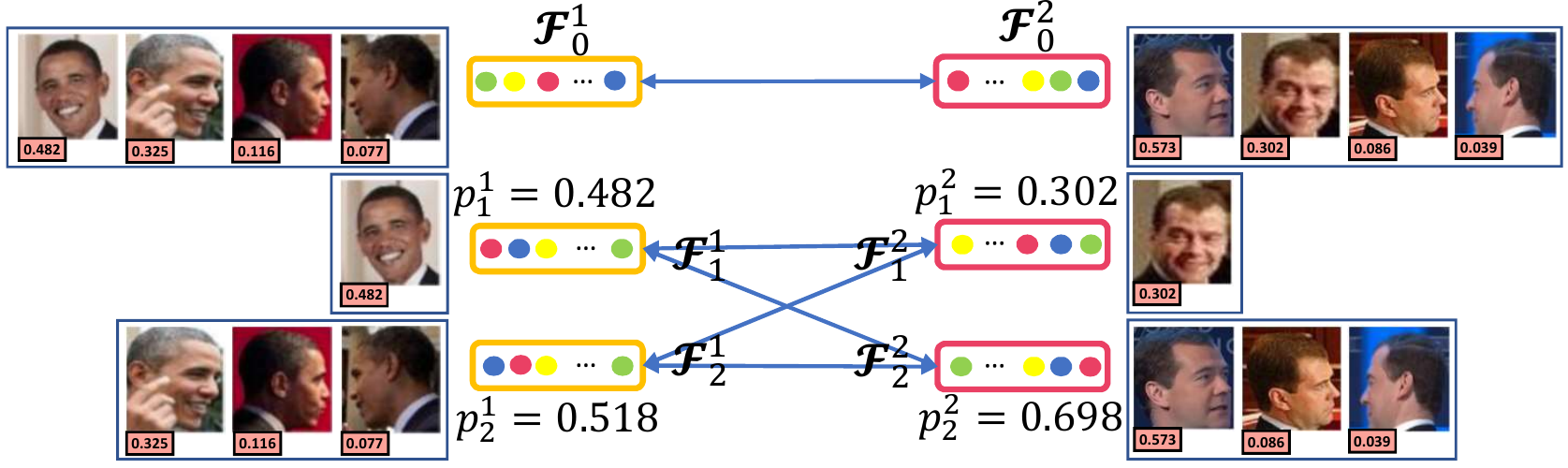}
\caption{ Illustration of the pose-guided representation scheme. }
\label{fig:example}
\end{figure}

\section{Numerical Experiments}

We evaluated the performance of the proposed method on three Set/video-based FR datasets: the IJB-A \cite{klare2015pushing}, YTF\cite{wolf2011face}, and Celebrity-1000\cite{liu2014toward}. To utilize the millions of available still images, we train our CNN embedding module separately. As in \cite{yang2017neural}, 3M face images from 50K identities are detected with the JDA \cite{chen2014joint} and aligned using the LBF \cite{ren2014face} method for our GoogleNet training. This part is fixed when we train the DAC module on each set/video face dataset. Benefiting from the highly-compact 128-d feature representation and the simple neural network of the DAC, the training time of our DAC(off) on IJB-A dataset with a single Xeon E5 v4 CPU is about 3 hours, the average testing time per each set-pair for verification is 62ms. We use Titan Xp for CNN processing.  

As our baseline methods, CNN$+$Mean L2 measures the average L2 distances of all image pairs of two sets, while the CNN$+$AvePool uses average pooling along each feature dimension for aggregation. The previous work NAN \cite{yang2017neural} uses the same CNN structure as our framework, but adopts a neural network module for independently quality assessment of each image. Therefore, NAN can be also regarded as our baseline. We refer the vanilla A2C as DAC(on), and use DAC(off) for the actor-critic with trust region-based experience replay scheme. The DAC(off)+PGR is the combination of the DAC(off) and PGR.

\subsection{Results on IJB-A dataset}

IJB-A \cite{klare2015pushing} is a face $verification$ and $identification$ dataset, containing images captured from unconstrained environments with wide variations of pose and imaging conditions. There are 500 identities with a total of 25,813 images (5,397 still images and 20,412 video frames sampled from 2,042 videos). A set of images for a particular identity is called a template. Each template can be a mixture of still images and sampled video frames. The number of images (or frames) in a template ranges from 1 to 190 with approximately 11.4 images and 4.2 videos per subject on average. It provides a ground truth bounding box for each face with 3 landmarks. There are 10 training and testing splits. Each split contains 333 training and 167 testing identities.

\begin{figure}[t!]
\centering
\begin{tabular}{cc}
\includegraphics[height=4.5cm]{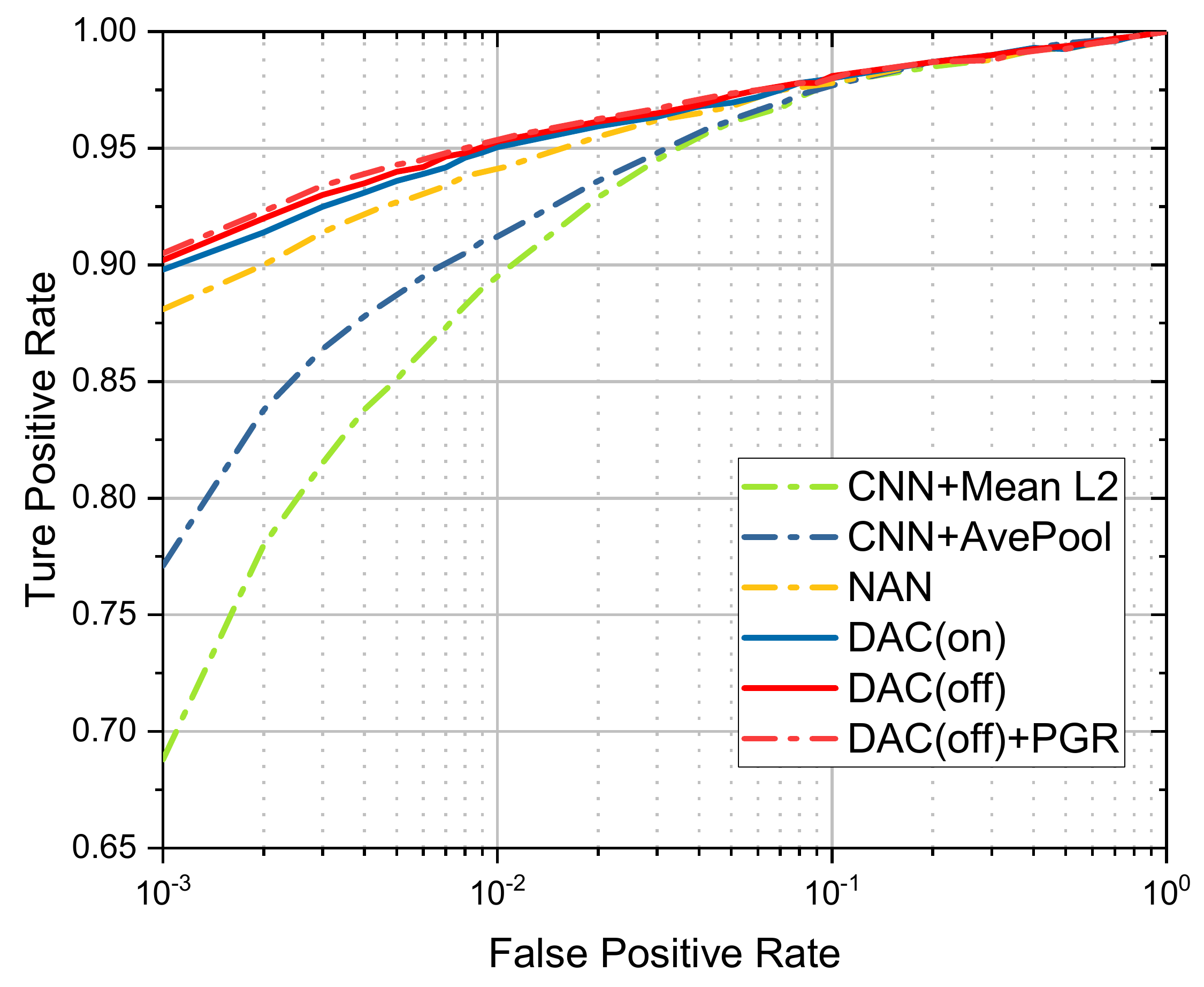}~~~~~~~&\includegraphics[height=4.5cm]{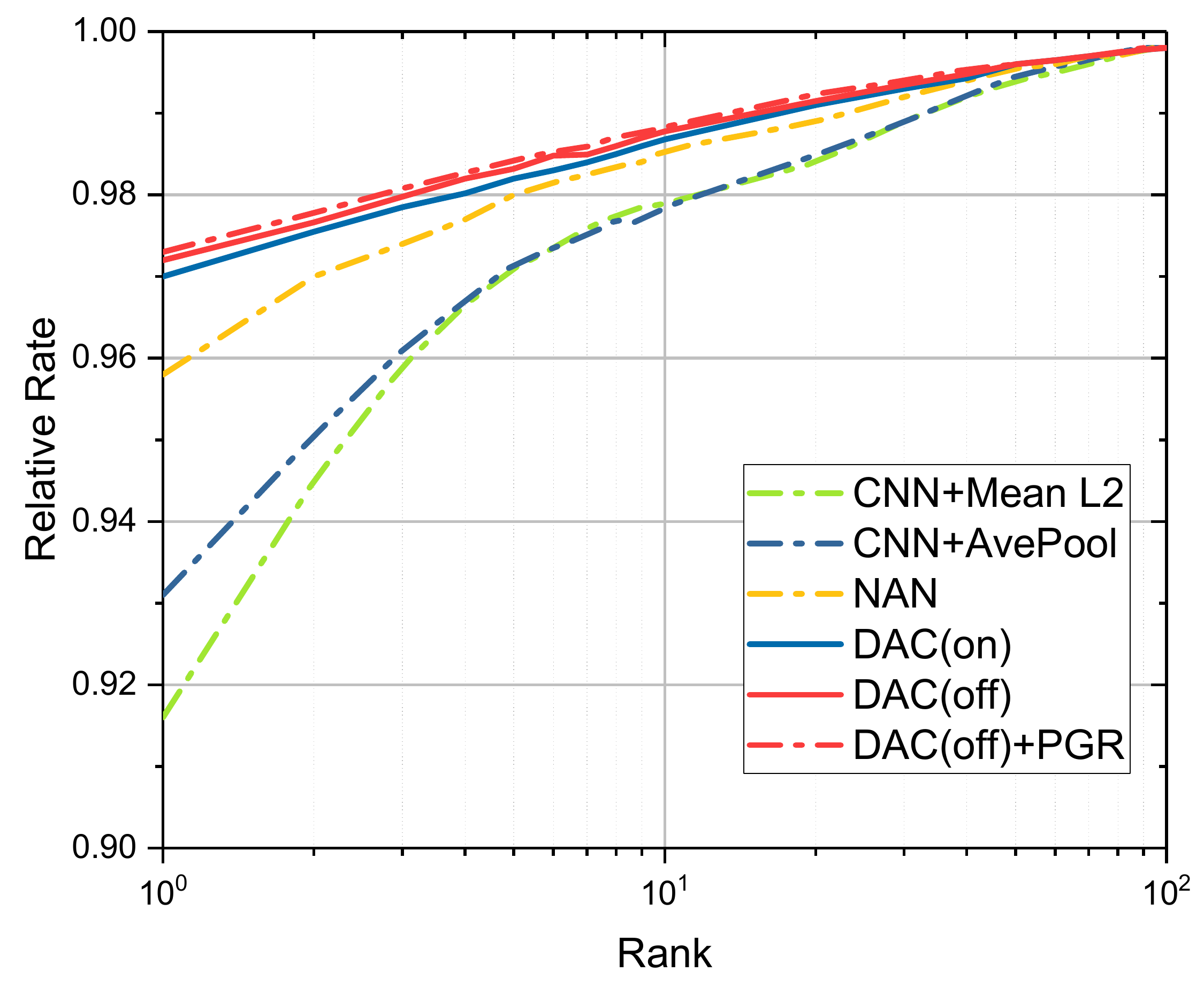}\\
(a)&~~~~~~(b)
\end{tabular}
\caption{Average ROC (Left) and CMC (Right) curves of the proposed method and its baselines on the IJB-A dataset over 10 splits.}
\label{fig:example}
\end{figure}

We compare the proposed framework with the existing methods on both face verification and identification following the standard evaluation protocol on IJB-A dataset. Metrics for the 1:1 compare task are evaluated using the receiver operating characteristics (ROC) curves in Fig. 5 (a). We also report the true accept rate (TAR) $vs.$ false positive rates (FAR) in Table 1. For the 1:N search task, the performance is evaluated in terms of a Cumulative Match Characteristics (CMC) curve as shown in Fig. 5 (b). It is an information retrieval metric, which plots identification rates corresponding to different ranks. A rank-$k$ identification rate is defined as the percentage of probe searches whose gallery match is returned with in the top-$k$ matches. The true positive identification rate (TPIR) $vs.$ false positive identification rate (FPIR) as well as the rank-1 accuracy are also reported in Table 1.

\begin{table}
\caption{Performance evaluation on the IJB-A dataset. For verification, the true accept rates (TAR) vs. false positive rates (FAR) are reported. For identification, the true positive identification rate (TPIR) vs. false positive identification rate (FPIR) and the Rank-1 accuracy are presented.}
\label{tab:different_nets}
\vspace{-5mm}
\begin{center}
\begin{tabular}{|c|c|c|c|c|c|}
    \hline
    \multirow{2}*{Method}& \multicolumn{2}{c|}{1:1 Verification TAR}& \multicolumn{3}{c|}{1:N Identification TPIR} \\ \cline{2-6}
    & FAR=0.01 & FAR=0.1 & FPIR=0.01 & FPIR=0.1 & Rank-1 \\ \hline \hline
    B-CNN\cite{chowdhury2016one} & - & - & 0.143$\pm$0.027 & 0.341$\pm$0.032 & 0.588$\pm$0.02\\ \hline
    LSFS\cite{wang2015face} & 0.733$\pm$0.034 & 0.895$\pm$0.013 & 0.383$\pm$0.063 & 0.613$\pm$0.032 & 0.820$\pm$0.024\\ \hline
    DCNN\cite{chen2015end}& 0.787$\pm$0.043 & 0.947$\pm$0.011 & - & - & 0.852$\pm$0.018\\ \hline
   Pose-model\cite{masi2016pose}& 0.826$\pm$0.018 & - & - & - & 0.840$\pm$0.012\\ \hline
   
   Masi~$et~al.$\cite{masi2016we} & 0.886 & - & - & - & 0.906\\ \hline
   Adaptation\cite{crosswhite2017template} & 0.939$\pm$0.013 & 0.979$\pm$0.004 & 0.774$\pm$0.049 & 0.882$\pm$0.016 & 0.928$\pm$0.010 \\ \hline
   QAN\cite{liu2017quality} & 0.942$\pm$0.015 & 0.980$\pm$0.006 & - & - & -\\ \hline
   NAN\cite{yang2017neural} & 0.941$\pm$0.008 & 0.978$\pm$0.003 & 0.817$\pm$0.041 & 0.917$\pm$0.009 & 0.958$\pm$0.005\\ \hline \hline
    DAC(on) & 0.951$\pm$0.014 & 0.980$\pm$0.016 & 0.852$\pm$0.048 & 0.931$\pm$0.012 & 0.970$\pm$0.011\\ \hline
    DAC(off) & 0.953$\pm$0.009 & \textbf{0.981}$\pm$\textbf{0.013}& 0.853$\pm$0.033 & 0.933$\pm$0.006& 0.972$\pm$0.012\\ \hline
    DAC(off)PGR &\footnotesize{\textbf{0.954}}$\pm$\textbf{0.01}& \textbf{0.981}$\pm$\textbf{0.008}& \textbf{0.855}$\pm$\textbf{0.042}&\textbf{0.934}$\pm$\textbf{0.009}&\textbf{0.973}$\pm$\textbf{0.011}\\ \hline
\end{tabular}
\end{center}
\end{table}

These results show that both the verification and the identification performance are largely improved compared to our baseline methods. The RL networks have learned to be robust to low-quality and redundant image. The DAC(on) outperforms the previous approaches in most of the operating points, showing that our representation is more discriminative than the weighted feature in \cite{yang2017neural,liu2017quality} without considering the inner-set dependency. The experience replay can further help the stabilization of our training and the state-of-the-art performance is achieved. Combining the off-policy DAC and pose-guide representation scheme also contributes to the final results in an efficient way.

\subsection{Results on YouTube Face dataset}

The YouTube Face (YTF) dataset \cite{wolf2011face} is a widely used video face $verification$ dataset, which contains 3,425 videos of 1,595 different subjects. In this dataset, there are many challenging videos, including amateur photography, occlusions, problematic lighting, pose and motion blur. The length of face videos in this dataset varies from 48 to 6,070 frames, and the average length of videos is 181.3 frames. In experiments, we follow the standard verification protocol as in \cite{yang2017neural,rao2017attention,rao2017learning}, which test our method for unconstrained face 1:1 verification with the given 5,000 video pairs. These pairs are equally divided into 10 splits, and each split has around 250 intra-personal pairs and 250 inter-personal pairs.

Table 2 presents the results of our DAC and previous methods. It can be seen that the DAC outperforms all the previous state-of-the-art methods following the setting that without fine-tuning the feature embedding module on YTF. Since this dataset has frontal face bias \cite{crosswhite2017template} and the face variations in this dataset are relatively small as shown in Fig. 2, we have not used the pose-guided representation scheme. It is obvious that the video sequences are redundant, considering the inner-video relationship does contribute to the improvement over \cite{yang2017neural}. The comparable performance with temporal representation-based methods suggests the DAC could be a potential substitute for RNN in some specific areas. Actually, the RNN itself is computationally expensive and sometimes difficult to train \cite{zhang2017towards}. We directly model the dependency in the feature-level, which is faster than the temporal representation of original images \cite{rao2017attention}, and more effective than the adversarial face generation-based method \cite{rao2017learning}.

\begin{table}[t!]
\caption{Comparisons of the average verification accuracy with the recently state-of-the-art results on the YTF dataset.$\dag$ fine-tuned the CNN model with YTF.
~~~~~~~~~~~~~~~~~~~~~~~~~~~~~~~~~~~~~~~~~~~~~~~~~~~~~~~~~~~~~~~~~~~~~~~~~~~~~~~~~~~~~~~~~~~~~~~~~~~~~~~~~~~~~~~~~~~~~~~~~~~~~~~~~~~~~~~~~~~~~~~~~~~~~~~~~~~~~~~~~~~~} 
\label{tab:different_nets}
\vspace{-5mm}
\begin{center}
\begin{tabular}{|c|c|c|c|}
   \hline
   Method&Accuracy&$\dag$Accuracy&Year\\ \hline \hline
   FaceNet\cite{schroff2015facenet}& 0.9512$\pm$0.0039&-&2015\\
   Deep FR\cite{parkhi2015deep}& 0.915& 0.973&2015\\
   CenterLoss\cite{wen2016discriminative}&0.949&-&2016\\ 
   TBE-CNN\cite{ding2017trunk}&0.9384$\pm$0.0032&0.9496$\pm$0.0031&2017\\
   TR\cite{rao2017attention}&0.9596$\pm$0.0059&0.9652$\pm$0.0054&2017\\
   NAN\cite{yang2017neural}&0.9572$\pm$0.0064&-&2017\\
   DAN\cite{rao2017learning}&0.9428$\pm$0.0069&-&2017\\ \hline \hline
   DAC(on)&0.9597$\pm$0.0041\\ \cline{1-2}
   DAC(off)&\textbf{0.9601}$\pm$\textbf{0.0048}\\ \cline{1-2}
\end{tabular}
\end{center}
\end{table}

It also indicates that DAC achieves a very competitive performance without highly-engineered CNN models. Note that the FaceNet \cite{schroff2015facenet}, NAN \cite{yang2017neural} also use the GoogleNet style structure. We show that DAC outperforms them on both the verification accuracy and the standard variation. The Deep FR, TBE-CNN and TR methods have additional fine-tuning of the CNN-model with YTF dataset, and the residual constitutional networks are used in TR. Considering our module-based structure, these advanced CNNs can be easily added on the DAC and boost its performance. We see that the DAC can generalizes well in video-based face verification datasets.

\subsection{Results on Celebrity-1000 dataset}

We then test our method on the Celebrity-1000 dataset \cite{liu2014toward}, which is designed for the unconstrained video-based face $identification$ problem. 2.4M frames from 159,726 face videos (about 15 frames per sequence) of 1,000 subjects are contained in this dataset. It is released with two standard evaluation protocols: open-set and closed-set. We follow the standard $1:N$ identification setting as in \cite{liu2014toward} and report the result of both protocols.

For the closed-set protocol, we use the softmax outputs from the reward network, and the subject with the maximum score as the result. Since the baseline methods do not have a multi-class prediction unit, we simply compare the L2 distance as in \cite{yang2017neural}. We present the results in Table 3, and show the CMC curves in Fig. 6 (a). With the help of end-to-end learning and large volume training data for CNN model, deep learning methods outperform \cite{liu2014toward,li2014eigen} by a large margin. It can be seen that the state-of-the-art is achieved by the DAC. We can also benefit from the experience replay to achieve improvements over the baselines.

\begin{figure}[t!]
\centering
\begin{tabular}{cc}
\includegraphics[height=4.5cm]{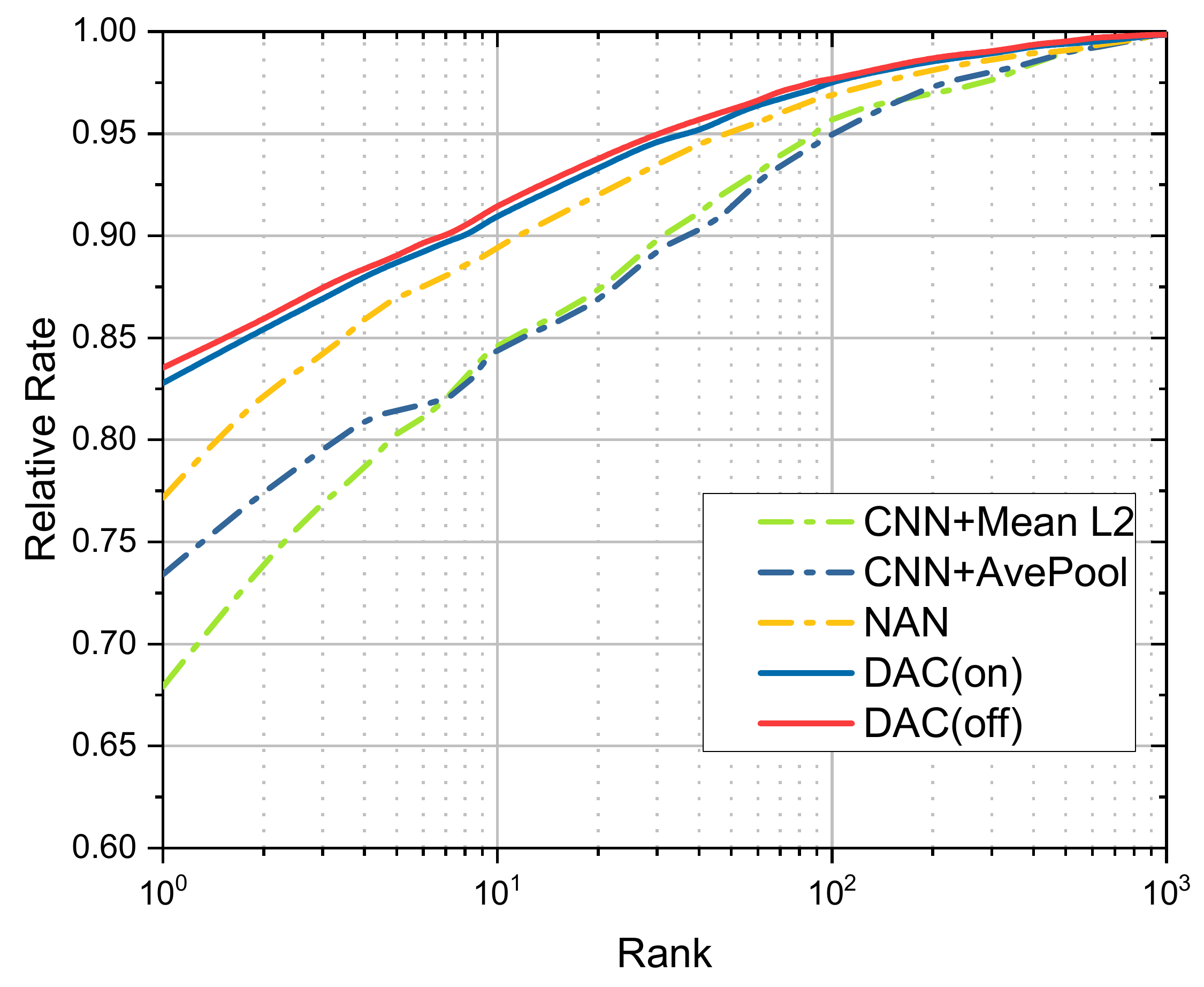}~~~&\includegraphics[height=4.5cm]{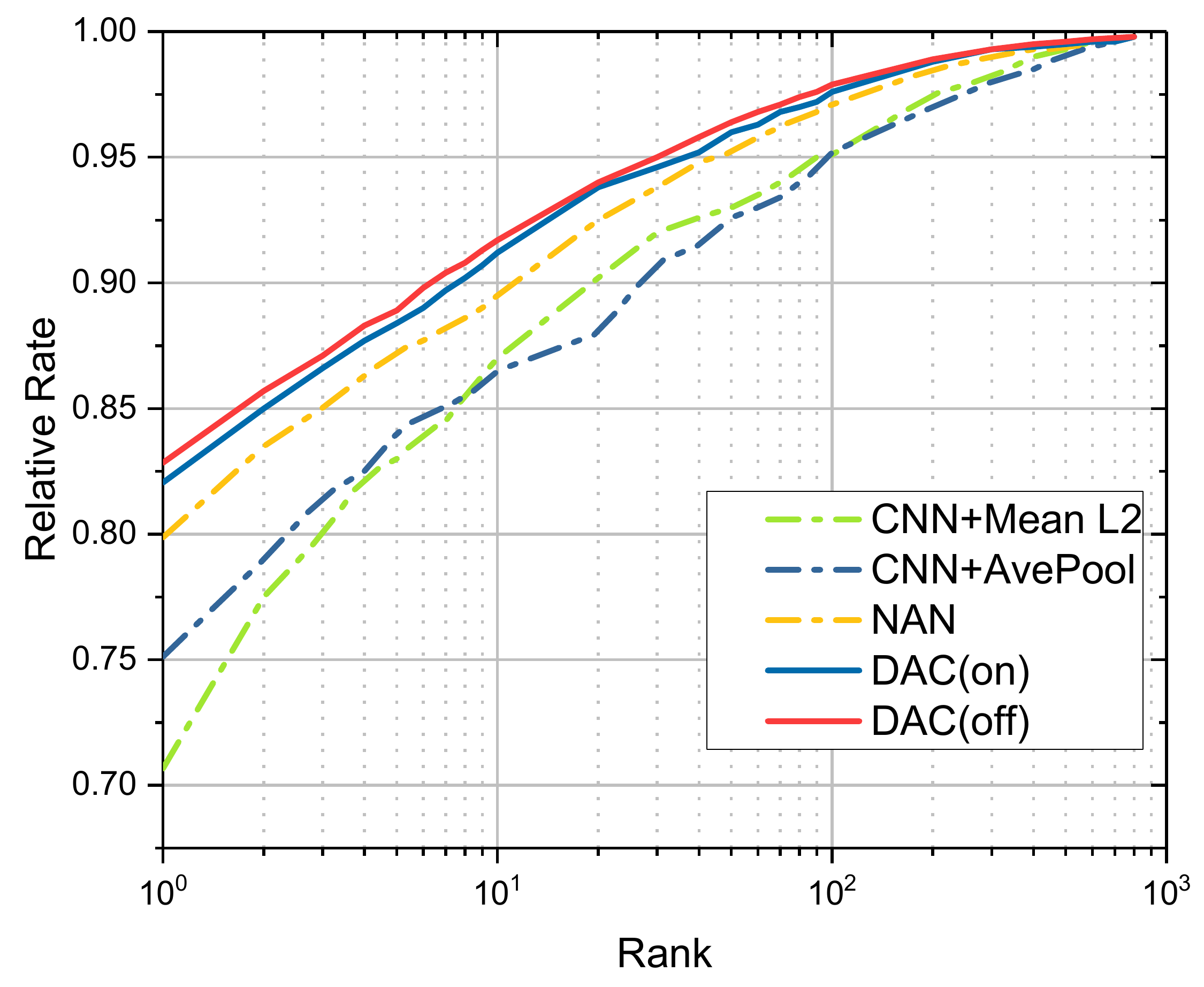}\\
(a)&~~~~~~(b)
\end{tabular}
\caption{
The CMC curves of different methods on Celebrity 1000. (a) Close-set tests on 1000 subjects, (b) Open-set tests on 800 subjects}.
\label{fig:example}
\end{figure}

\begin{table}[t!]
\caption{Identification performance (rank-1 accuracies), on the Celebrity-1000 dataset for the closed-set tests (left) and open-set tests (right).
~~~~~~~~~~~~~~~~~~~~~~~~~~~~~~~~~~~~~~~~~~~~~~~~~~~~~~~~~~~~~~~~~~~~~~~~~~~~~~~~~~~~~~~~~~~~~~~~~~~~~~~~~~~~~~~~~~~~~~~~~~~~~~~~~~~~~~~~~~~~~~~~~~~~~~~~~~~~~~~~~~~~} 
\label{tab:different_nets}
\vspace{-5mm}
\begin{center}
\begin{tabular}{|c|c|c|c|c|c|c|c|c|c|c|}
    \cline{1-1} \cline{3-6} \cline{8-11}
    \multirow{2}*{Method}&~& \multicolumn{4}{c|}{Number of Subjects($closed$)}&~& \multicolumn{4}{c|}{Number of subjects($open$)}\\ \cline{3-6} \cline{8-11}

      &~& 100 & 200 & 500 & 1000 &~& 100 & 200 & 500 & 800\\  \hline \hline
   MTJSR\cite{liu2014toward}&~&0.506&0.408&0.3546&0.3004&~&0.4612&0.3984&0.3751&0.3350\\ \cline{1-1} \cline{3-6} \cline{8-11}
   Eigen-PEP\cite{li2014eigen}&~&0.506&0.4502&0.3997&0.3194&~&0.5155&0.4615&0.4233&0.2590\\ \cline{1-1} \cline{3-6} \cline{8-11}
   CNN+Mean L2&~&0.8526&0.7759&0.7457&0.6791&~&0.8488&0.7988&0.7676&0.7067\\ \cline{1-1} \cline{3-6} \cline{8-11}
   CNN+AvePool&~&0.8446&0.7893&0.7768&0.7341&~&0.8411&0.7909&0.7840&0.7512\\ \cline{1-1} \cline{3-6} \cline{8-11}
   NAN\cite{yang2017neural}&~&0.9044&0.8333&0.8227&0.7717&~&0.8876&0.8521&0.8274&0.7987\\ \hline \hline
   DAC(on)&~&0.9125&0.8722&0.8475&0.8278&~&0.8986&0.8706&0.8395&0.8205\\ \cline{1-1} \cline{3-6} \cline{8-11}
   DAC(off)&~&\textbf{0.9137}&\textbf{0.8783}&\textbf{0.8523}&\textbf{0.8353}&~&\textbf{0.9004}&\textbf{0.8715}&\textbf{0.8428}&\textbf{0.8264}\\ \cline{1-1} \cline{3-6} \cline{8-11}

\end{tabular}
\end{center}
\end{table}

For the open-set testing, we take multiple image sequences of each gallery subject to extract a highly compact feature representation as in NAN \cite{yang2017neural}. Then the open-set identification is performed by comparing the L2 distance of the aggregated probe and gallery representations. Fig. 6 (b) and Table 3 show the results of different methods in our experiments. We see that our proposed methods outperform the previous methods again, which clearly shows that DAC is effective and robust.

\section{Conclusions}
We have introduced the actor-critic RL for visual recognition problem. We cast the inner-set dependency modeling to a MDP, and train an agent DAC to make attention control for each image in each step. The PGR scheme well balances the computation cost and information utilization. Although we only explore their ability in set/video-based face recognition tasks, we believe it is a general and practicable methodology that could be easily applied to other problems, such as Re-ID, action recognition and event detection $etc$.

\section{Acknowledgement}
This work was supported in part by the National Key R\&D Plan 2016YFB0501003, Hong Kong Government General Research Fund GRF 152202/14E, PolyU Central Research Grant G-YBJW, Youth Innovation Promotion Association, CAS (2017264), Innovative Foundation of CIOMP, CAS (Y586320150).

%
%
%
%
\bibliographystyle{splncs}
\bibliography{egbib}
\end{document}